\pdfoutput=1

\documentclass[11pt]{article}

\usepackage[]{acl}

\usepackage{times}
\usepackage{latexsym}
\usepackage{amsmath,amssymb,amsthm}
\usepackage{mathabx,mathrsfs}
\usepackage{relsize}
\usepackage[T1]{fontenc}

\usepackage[utf8]{inputenc}
\usepackage{microtype}
\usepackage{CJKutf8}
\usepackage{bm}
\usepackage{booktabs}
\usepackage{caption}
\usepackage{subcaption}
\usepackage{pifont}%
\newcommand{\cmark}{\ding{51}}%
\newcommand{\xmark}{\ding{55}}%
\usepackage{multirow}
\usepackage{hyperref}
\usepackage{microtype}
\usepackage{tikz}
\usepackage{pgfplots}
\usepackage{CJKutf8}
\usepackage{tikz-dependency}
\usepackage{graphicx}
\usepackage{natbib}
\usepackage{tikz-qtree}
\usepgflibrary{arrows.meta}
\usepgfplotslibrary{fillbetween}
\usetikzlibrary{fit, calc, decorations.pathreplacing, positioning, shapes.geometric, backgrounds}
\usepackage{float}

\usepackage{graphicx}

\newcommand\blfootnote[1]{%
  \begingroup
  \renewcommand\thefootnote{}\footnote{#1}%
  \addtocounter{footnote}{-1}%
  \endgroup
}

\title{SynGEC: Syntax-Enhanced Grammatical Error Correction \\ with a Tailored GEC-Oriented Parser}

\author{Yue Zhang$^{1}$, {\bf Bo Zhang$^2$}, Zhenghua Li$^{1*}$, Zuyi Bao$^2$,   {\bf Chen Li$^2$}, {\bf Min Zhang$^1$} \\
        $^1$Institute of Artificial Intelligence, School of Computer Science and Technology, \\
Soochow University, China; $^2$DAMO Academy, Alibaba Group, China\\
\texttt{$^1$yzhang21@stu.suda.edu.cn}, \texttt{$^1$\{zhli13,minzhang\}@suda.edu.cn}\\\texttt{$^2$\{klayzhang.zb,zuyi.bzy,puji.lc\}@alibaba-inc.com}}

\begin{document}
\maketitle
\begin{abstract}

This work proposes a syntax-enhanced grammatical error correction (GEC) approach named SynGEC that effectively incorporates \blfootnote{$^*$ Corresponding author.} dependency syntactic information into the encoder part of GEC models.\footnote{Although this work focuses on the dependency syntax structure, SynGEC can also be extended to the constituency syntax structure straightforwardly.} The key challenge for this idea is 
that off-the-shelf parsers %
are unreliable when processing 
ungrammatical sentences. 
To confront this challenge, we propose to build a tailored GEC-oriented parser (GOPar) using parallel GEC training data as a pivot. 
First, we design an extended syntax representation scheme that allows us to represent both grammatical errors and syntax in a unified tree structure.
Then, we obtain parse trees of the source incorrect sentences by projecting trees of the target correct sentences.
Finally, we train GOPar with such projected trees. 
For GEC, we employ the graph convolution network to encode source-side syntactic information produced by GOPar, and fuse them with the outputs of the Transformer encoder.
Experiments on mainstream English and Chinese GEC datasets show that our proposed SynGEC approach consistently and substantially outperforms strong baselines and achieves competitive performance. Our code and data are all publicly available at \url{https://github.com/HillZhang1999/SynGEC}.

\end{abstract}

\section{Introduction}
\label{intro}
Given an ungrammatical sentence, the grammatical error correction (GEC) task aims to produce a grammatical target sentence with the intended meaning \citep{grundkiewicz2020crash,wang2021comprehensive}. %
Recent mainstream approaches treat GEC as a monolingual machine translation (MT) task  \citep{yuan2016grammatical,junczys2018approaching}. 
Standard encoder-decoder based MT models,  e.g., Transformer \citep{vaswani2017attention}, have emerged as a dominant paradigm and achieved  state-of-the-art (SOTA) results on various GEC benchmarks \citep{rothe2021recipe, stahlberg2021synthetic, sun2022unified, zhang2022mucgec}. 
Despite their impressive achievements, most work treats the input sentence as a sequence of tokens, without explicitly exploiting syntactic or semantic information.

Compared with MT, GEC has two peculiarities that directly motivate this work.
First, the training data for GEC models is much less abundant, which may be alleviated by incorporating linguistic structure knowledge like syntax. 
As shown in Table \ref{tab:dataset} and Table \ref{tab:chinese:dataset}, the English and Chinese GEC tasks only have about 126K and 157K high-quality labelled source/target sentence pairs for training, if not  considering the highly noisy crowd-annotated Lang8 data  \citep{mita2020self}. 
Second, according to our preliminary observation, many errors in ungrammatical sentences are intrinsically correlated with syntactic information. 
For example, errors like inconsistency in tense or singular-vs-plural forms can be better detected and corrected with the help of long-range syntactic dependencies.  

\definecolor{brickred}{HTML}{b92622}
\definecolor{midnightblue}{HTML}{005c7f}
\definecolor{salmon}{HTML}{f1958d}
\definecolor{burntorange}{HTML}{f19249}
\definecolor{junglegreen}{HTML}{4dae9d}
\definecolor{forestgreen}{HTML}{499c5e}
\definecolor{pinegreen}{HTML}{3d8a75}
\definecolor{seagreen}{HTML}{6bc1a2}
\definecolor{limegreen}{HTML}{97c65a}
\newcommand{\white}[1]{\textcolor{white}{#1}}
\newcommand{\brickred}[1]{\textcolor{brickred}{#1}}
\newcommand{\midnightblue}[1]{\textcolor{midnightblue}{#1}}
\newcommand{\salmon}[1]{\textcolor{salmon}{#1}}
\newcommand{\junglegreen}[1]{\textcolor{junglegreen}{#1}}
\newcommand{\forestgreen}[1]{\textcolor{forestgreen}{#1}}
\newcommand{\pinegreen}[1]{\textcolor{pinegreen}{#1}}
\newcommand{\seagreen}[1]{\textcolor{seagreen}{#1}}

\begin{figure*}[th!]
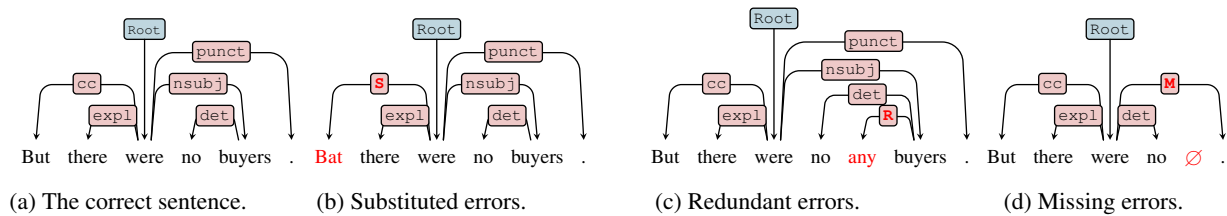

  \begin{subfigure}[b]{0.2\textwidth}
    \centering
    \scalebox{0.62}{
        \begin{dependency}
          \begin{deptext}[column sep=0.2cm,font=\small]
            \large{But} \& \large{there} \& \large{were} \& \large{no} \& \large{buyers} \& \large{.} \\ 
          \end{deptext}
          \Large{
          \deproot[edge vertical padding=0.6ex, edge height=12ex, label style={fill=midnightblue!25, thick}, edge style={thick}]{3}{\texttt{\large{Root}}}
          \depedge[edge vertical padding=0.6ex, edge height=5.2ex, label style={fill=brickred!25, thick}, thick]{3}{1}{\texttt{cc}}
          \depedge[edge vertical padding=0.6ex, edge height=2.2ex, label style={fill=brickred!25, thick}, thick]{3}{2}{\texttt{expl}}
            \depedge[edge vertical padding=0.6ex, edge height=5.2ex, label style={fill=brickred!25, thick}, thick]{3}{5}{\texttt{nsubj}}
          \depedge[edge vertical padding=0.6ex, edge height=8.2ex, label style={fill=brickred!25, thick}, thick]{3}{6}{\texttt{punct}}
          \depedge[edge vertical padding=0.6ex, edge height=2.2ex, label style={fill=brickred!25, thick}, thick]{5}{4}{\texttt{det}}}
    
        \end{dependency}
    }
    
    \caption{The correct sentence.}
    \label{fig:origin-sent}
  \end{subfigure}
  \hfill
   \begin{subfigure}[b]{0.2\textwidth}
    \centering
    \scalebox{0.62}{
    
    \begin{dependency}
      \begin{deptext}[column sep=0.2cm,font=\small]
        \large{\textcolor{red}{Bat}} \& \large{there} \& \large{were} \& \large{no} \& \large{buyers} \& \large{.} \\ 
      \end{deptext}
      \Large{
      \deproot[edge vertical padding=0.6ex, edge height=12ex, label style={fill=midnightblue!25, thick}, edge style={thick}]{3}{\texttt{Root}}
      \depedge[edge vertical padding=0.6ex, edge height=5.2ex, label style={fill=brickred!25, thick}, thick]{3}{1}{\textcolor{red}{\texttt{\textbf{S}}}}
      \depedge[edge vertical padding=0.6ex, edge height=2.2ex, label style={fill=brickred!25, thick}, thick]{3}{2}{\texttt{expl}}
        \depedge[edge vertical padding=0.6ex, edge height=5.2ex, label style={fill=brickred!25, thick}, thick]{3}{5}{\texttt{nsubj}}
      \depedge[edge vertical padding=0.6ex, edge height=8.2ex, label style={fill=brickred!25, thick}, thick]{3}{6}{\texttt{punct}}
      \depedge[edge vertical padding=0.6ex, edge height=2.2ex, label style={fill=brickred!25, thick}, thick]{5}{4}{\texttt{det}}
      }

    \end{dependency}
    }
    \caption{Substituted errors.}
    \label{fig:s-error}
  \end{subfigure}
  \hfill
  \hfill
   \begin{subfigure}[b]{0.2\textwidth}
    \centering
    \scalebox{0.62}{
    \begin{dependency}
      \begin{deptext}[column sep=0.2cm,font=\small]
        \large{But} \& \large{there} \& \large{were} \& \large{no} \& \textcolor{red}{\large{any}} \& \large{buyers} \& \large{.} \\ 
      \end{deptext}
      \Large{
      \deproot[edge vertical padding=0.6ex, edge height=13ex, label style={fill=midnightblue!25, thick}, edge style={thick}]{3}{\texttt{Root}}
      \depedge[edge vertical padding=0.6ex, edge height=5.2ex, label style={fill=brickred!25, thick}, thick]{3}{1}{\texttt{cc}}
      \depedge[edge vertical padding=0.6ex, edge height=2.2ex, label style={fill=brickred!25, thick}, thick]{3}{2}{\texttt{expl}}
        \depedge[edge vertical padding=0.6ex, edge height=6.5ex, label style={fill=brickred!25, thick}, thick]{3}{6}{\texttt{nsubj}}
      \depedge[edge vertical padding=0.6ex, edge height=9.2ex, label style={fill=brickred!25, thick}, thick]{3}{7}{\texttt{punct}}
      \depedge[edge vertical padding=0.6ex, edge height=4.2ex, label style={fill=brickred!25, thick}, thick]{6}{4}{\texttt{det}}
      \depedge[edge vertical padding=0.6ex, edge height=2.2ex, label style={fill=brickred!25, thick}, thick]{6}{5}{\textbf{\textcolor{red}{\texttt{R}}}}
        }
    \end{dependency}
    }
    \caption{Redundant errors.}
    \label{fig:r-error}
  \end{subfigure}
  \hfill
  \hfill
   \begin{subfigure}[b]{0.2\textwidth}
    \centering
    \scalebox{0.62}{
    \begin{dependency}
      \begin{deptext}[column sep=0.2cm,font=\small]
        \large{But} \& \large{there} \& \large{were} \& \large{no} \& \textcolor{red}{\large{$\emptyset$}} \& \large{.} \\ 
      \end{deptext}
      \Large{
      \deproot[edge vertical padding=0.6ex, edge height=12ex, label style={fill=midnightblue!25, thick}, edge style={thick}]{3}{\texttt{Root}}
      \depedge[edge vertical padding=0.6ex, edge height=5.2ex, label style={fill=brickred!25, thick}, thick]{3}{1}{\texttt{cc}}
      \depedge[edge vertical padding=0.6ex, edge height=2.2ex, label style={fill=brickred!25, thick}, thick]{3}{2}{\texttt{expl}}
      \depedge[edge vertical padding=0.6ex, edge height=5.2ex, label style={fill=brickred!25, thick}, thick]{3}{6}{\textcolor{red}{\texttt{\textbf{M}}}}
      \depedge[edge vertical padding=0.6ex, edge height=2.2ex, label style={fill=brickred!25, thick}, thick]{3}{4}{\texttt{det}}
        }
    \end{dependency}
    }
    \caption{Missing errors.}
    \label{fig:m-error}
  \end{subfigure}
  \hfill
  \caption{Illustration of our extended syntax representation scheme. $\emptyset$ denotes the missing word.}
  \label{fig:scheme-three-error-types}
\end{figure*}

In this paper, we propose SynGEC, an approach that can effectively inject the syntactic structure of the input sentence into the encoder part of GEC models. 
The critical challenge here is that 
off-the-shelf parsers are unreliable when handling ungrammatical sentences. 
On the one hand, 
off-the-shelf parsers are trained on clean treebanks that only consist of  grammatical sentences. When parsing ungrammatical sentences, their performance may sharply degrade due to the input mismatch. 
On the other hand, mainstream syntax representation schemes, adopted by existing treebanks, do not cover the non-canonical structures arising from grammatical errors.
In consequence, under such schemes, it is sometimes difficult to find a plausible syntactic tree to properly parse an ungrammatical sentence (e.g., the sentence in Figure \ref{fig:m-error}).

Indeed, there have been several prior works that try to improve syntactic parsing for ungrammatical texts by annotating data  \citep{dickinson2009dependency,DBLP:conf/acl/BerzakKSWLMGK16,nagata2016phrase}. 
However, these works do not extend 
existing syntax representation schemes to accommodate errors, which means that they make little change on %
the original syntactic label sets. 
Besides, manual annotation is expensive and time-consuming, so their annotated treebanks for ungrammatical sentences are of a relatively small scale. 

To confront the challenge of unreliable performance of off-the-shelf parsers on ungrammatical sentences, 
we propose to train a tailored GEC-oriented parser (GOPar). The basic idea is to utilize parallel source/target sentence pairs in the GEC training data. First, we parse the target correct sentences using a vanilla off-the-shelf parser. Then, we construct the tree for the source incorrect sentences via tree projection. 
To accommodate grammatical errors, we propose an extended syntax representation scheme based on several straightforward rules, which allows us to represent both grammatical errors and syntax in a unified tree structure. 
Finally, we train GOPar directly on the automatically constructed trees of the source incorrect sentences in the GEC training data. 
During both GEC training and evaluation procedures, GOPar is used to generate syntactic information for the input sentences.

To incorporate syntactic information provided by GOPar, we cascade several label-aware graph convolutional network (GCN) layers  \citep{DBLP:conf/iclr/KipfW17, DBLP:conf/acl/ZhangZWLZ20} above the encoder of our baseline Transformer-based GEC model. 
We conduct experiments on two widely-used English GEC evaluation datasets, i.e., CoNLL-14 \citep{ng2014conll} and BEA-19 \citep{bryant2019bea}, and two Chinese GEC evaluation datasets, i.e., NLPCC-18 \cite{zhao2018overview} and MuCGEC \citep{zhang2022mucgec}. Extensive experimental results and in-depth analyses show that our 
SynGEC approach achieves consistent and substantial  improvement on all datasets, even when the baseline model is enhanced with large pre-trained language models (PLMs) like BART \citep{lewis2020bart}, and outperforming previous SOTA systems under comparable settings.

\section{Our GEC-Oriented Parser}
\label{sec:parsing:data}

This section describes our tailored GOPar, a dependency parser that is more competent in parsing ungrammatical sentences than off-the-shelf parsers.

\subsection{Extended Syntax Representation Scheme} 
\label{GOPar Scheme}  

The standard scheme for representing dependency syntax is originally designed for grammatical sentences, 
 and thus may not cover many non-canonical structures in grammatically erroneous sentences. 
Therefore, to obtain a tailored parser, our first task is to extend the syntax representation scheme and, more specifically, to design a complementary set of rules to handle different grammatical mistakes. 
With this scheme, we can directly use a unified tree structure to represent both grammatical errors and syntactic information. 

\begin{figure*}[ht!]
\centering
\includegraphics[scale=0.45]{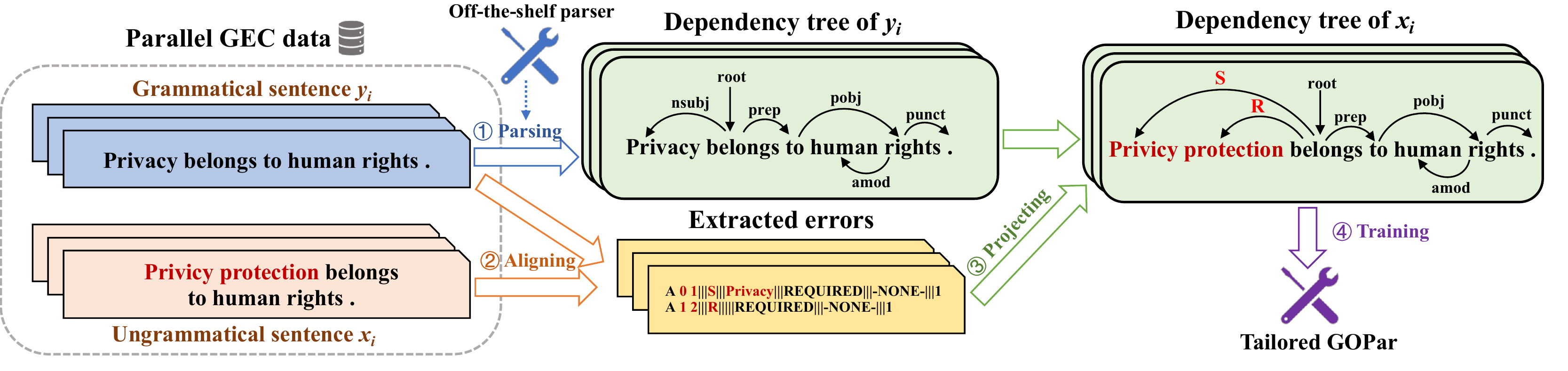}
\caption{The workflow for obtaining our tailored  GOPar. 
}
\label{fig:tree}
\end{figure*}

As shown in Figure \ref{fig:scheme-three-error-types}, we propose a light-weight extended scheme based on several straightforward rules, 
corresponding to the three types of grammatical errors, i.e., \textit{substituted}, \textit{redundant} and \textit{missing} \citep{bryant2017automatic}.\footnote{
We treat %
\textit{word-order} errors as the combination of redundant and missing errors in this work. 
} 
In this work, we use the Stanford Dependencies Scheme v3.3.0 \cite{DBLP:conf/iclr/DozatM17} as the basic scheme. 
The rules are designed in such a way that 
we make as few adjustments as possible to the  syntactic tree of the target correct sentence during tree projection. 
Correspondingly, we add three labels into the original syntactic label set, i.e., ``S'', ``R'' and ``M'', to capture three kinds of errors.  
Since such categorization is also adopted in the grammatical error detection (GED) task \citep{DBLP:conf/emnlp/YuanTD021}, we refer to them as GED labels.

\begin{itemize}
    \item \textbf{Substituted errors (S)} include spelling errors, tense errors, singular/plural inconsistency errors, etc. For simplicity, we do not consider such fine-grained categories, and directly use a single ``S'' label to indicate that the word should be replaced by another one, as shown in Figure \ref{fig:s-error}.
    \item \textbf{Redundant errors (R)} mean that some words should be deleted. For each redundant word, we let it depend on its right-side adjacent word, with a label ``R'',\footnote{Even if the right-side adjacent word is redundant.} as shown in Figure \ref{fig:r-error}. When the redundant word is at the end of the sentence, we instead let it depend on its left-side adjacent word. 
    \item \textbf{Missing errors (M)} mean that some words should be inserted. For each missing word, we assign a label ``M'' to the incoming arc of its right-side adjacent word, as shown in Figure \ref{fig:m-error}. When the missing word is at the end of the sentence, we  keep the original tree unchanged. If several consecutive words are missing, the structure remains the same as when a single word is  missing.  Moreover, since a missing word may have children in the tree of the correct sentence, we let them depend on the head word of the missing word, without changing their  syntactic labels. 
\end{itemize}

\textbf{Limitation discussion.} %
Our extended scheme may 
encounter problems when different types of errors occur consecutively. %
Taking ``But was no buyers'' as an example, we need to replace ``was'' with ``were'' and then insert ``there'' before ``were'' at the same time. 
Therefore, according to our rules, the label of the incoming arc of ``was'' can be either ``S'' or ``M'', leading to a label conflict. 
To decide a unique label, we simply define a priority order: ``S''$>$``R''$>$``M'.
Overall, the current version of our scheme is imperfect, and there are still many points that can be improved. For example, when confronting substituted and redundant errors, some original labels will be overwritten by the GED label ``S'' and ``M'', which may cause the loss of some valuable information. One possible solution is to combine GED and syntax labels and use joint labels like ``S-Root'', ``M-Subj'', etc. We leave such extensions of our scheme as future work.

\subsection{Training GOPar} %
\label{GOPar Data}  

With the extended syntax representation scheme, we propose to train our tailored GOPar by using the parallel GEC training data $D=\{(x_i, y_i)\}$ as a pivot. 
The major goal is to automatically generate high-quality parse trees for large-scale sentences with realistic grammatical errors, and use them to train a parser suitable for parsing ungrammatical sentences. 
Figure \ref{fig:tree} illustrates the workflow, consisting of the following four steps.

First, we use an off-the-shelf parser to parse the target correct sentences (i.e., $y_i$) of the GEC training data. The off-the-shelf parser can produce reliable parse trees for target-side sentences since they are (ideally) free from grammatical errors.

Second, we employ ERRANT \citep{bryant2017automatic}\footnote{\url{https://github.com/chrisjbryant/errant}} 
to extract all grammatical errors in the source incorrect sentence (i.e., $x_i$) according to the alignments  between $x_i$ and $y_i$. The errors extracted by ERRANT mainly contain 3 parts: the start and end positions of errors in source sentences, the corresponding corrections, and the error types.

Third, we construct the tree of $x_i$ by projecting the target-side tree of $y_i$ to the source side. For words that are not related to any errors, dependencies and labels are directly copied; for those related to errors,  dependencies and labels are assigned according to the rules introduced in Section \ref{GOPar Scheme}.

Fourth,  with constructed parse trees for all source-side sentences in $D$,  we then use them as a treebank to train our tailored GOPar.

\textbf{An alternative way} to build  GOPar is directly utilizing manually labeled treebanks. 
Since existing treebanks only contain grammatical sentences, we can inject synthetic errors based on rules or back-translation models \citep{foster2008adapting,cahill2015parsing}. 
Then, we can produce parse trees for ungrammatical sentences analogously through the above second and third steps. 
However, our preliminary experiments show that GOPar built in this way is much inferior and can only 
slightly improves our baseline GEC model.  
We suspect that the reasons are two-fold. On the one hand, there is a considerable gap between synthetic and real grammatical errors; on the other hand, the generated data is not enough to train GOPar adequately due to the limited scale of existing treebanks, as GOPar needs to learn to accommodate multifarious errors.

\section{The DepGCN-based GEC Model}
\begin{figure}[t!]
\centering
\includegraphics[scale=0.45]{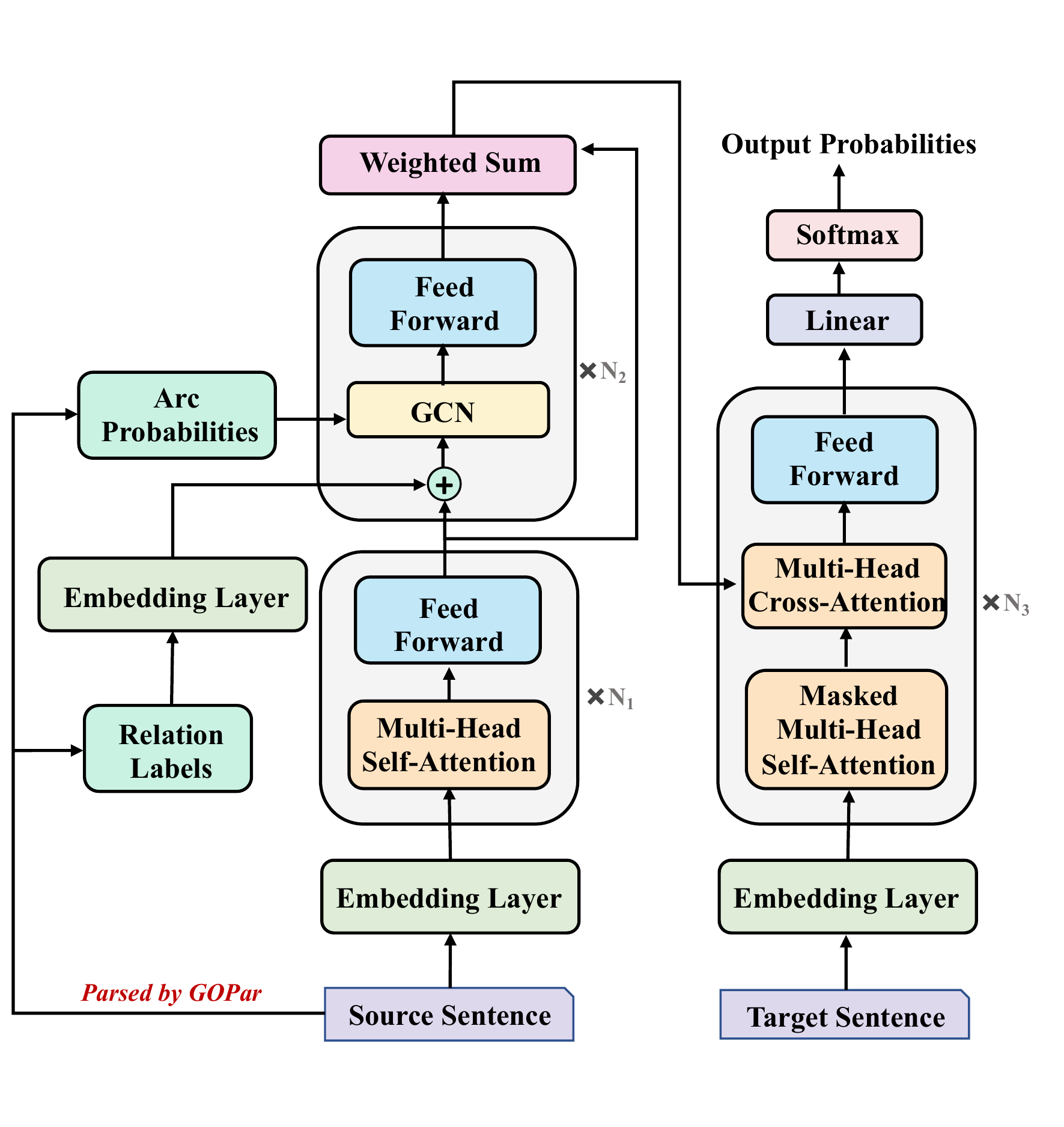}
\caption{Overview of our DepGCN-based GEC model. The operation $\oplus$ denotes vector concatenation. $N_1$, $N_2$ and $N_3$ denote the number of identical Transformer encoder blocks, DepGCN blocks and Transformer decoder blocks, respectively. We also employ a residual connection \citep{he2016deep} followed by layer normalization \citep{DBLP:journals/corr/BaKH16} around each sub-layer.}
\label{fig:model}
\end{figure}
This section describes our DepGCN-based GEC model, whose architecture is shown in Figure \ref{fig:model}.
We adopt GCN \citep{DBLP:conf/iclr/KipfW17} to encode the dependency syntax trees of the source sentence. Then, we feed the encoded syntactic information into a Transformer-based GEC model.

\subsection{Transformer Backbone}
We model GEC as a sequence-to-sequence task and employ the commonly used Transformer model \citep{vaswani2017attention} as the backbone. The Transformer is composed of an encoder and a decoder. The encoder utilizes the multi-head self-attention mechanism to get the contextualized representations of each token in the source sentence. The decoder has a similar architecture while additionally containing a masked multi-head self-attention module to model the generated token information.

During training, the objective function is to minimize the teacher forcing negative log-likelihood loss \citep{williams1989learning}, formally:
\begin{equation}
\begin{aligned}
\mathcal{L}(\theta) &=- \log (P(y \mid x;\theta))\\
                &=- \log \sum_{t=1}^{n}(P(y_t \mid y_{<t};x;\theta))
\end{aligned}
\end{equation}
where $\theta$ is trainable model parameters, $x$ is the source sentence,  $y=\{y_1,y_2,...,y_n\}$ is the ground-truth target sentence with $n$ tokens, and  $y_{<t}=\{y_1,y_2,...,y_{t-1}\}$ is the tokens visible in $t$-th training time step.

During inference,  we utilize beam search decoding \citep{DBLP:conf/emnlp/WisemanR16} to find an optimal sequence $y^*$ by maximizing the conditional probability $P(y^* \mid x;\theta)$.  

Previous work shows that PLMs, e.g., BART \citep{lewis2020bart} and T5 \citep{raffel2020exploring}, can improve GEC performance over training from scratch by large margins \citep{rothe2021recipe,sun2022unified}. 
In this work, we further use BART to build a stronger baseline, since it shares the same model architecture with our Transformer backbone. 
Specifically, we use the BART parameters to initialize our Transformer backbone and then continue training on GEC training data. More details are discussed in Section \ref{sec:english:exp} and Section \ref{sec:chinese:exp}. 

\subsection{Dependency GCN (DepGCN)}
We employ DepGCN \cite{DBLP:conf/acl/ZhangLZ20} to encode dependency syntax information.
The DepGCN module stacks several identical blocks, and each block is composed of a GCN sub-layer and a feed-forward sub-layer.

For the GCN sub-layer, we introduce the information of the dependency arcs and dependency labels simultaneously. We compute the output $\mathbf{h}_i^{(l)}$ of $l$-th GCN at the $i$-th token as:
\begin{equation}
\label{eq-GCN}
    \mathbf{h}_i^{(l)}=\mbox{ReLU}(\sum^n_{j=1}A_{ij}W^{(l)}(\mathbf{h}_j^{l-1} \oplus \mathbf{e}^{(i,j)} +\mathbf{b}^{(l)})
\end{equation}
where $A\in\mathbb{R}^{n \times n}$ denotes the adjacency matrix, 
$\mathbf{e}^{{(i,j)}} \in \mathbb{R}^{d}$ is the embedding of the dependency label between word $w_i$ and word $w_j$, $W\in\mathbb{R}^{d \times 2d}$ and $\mathbf{b}\in\mathbb{R}^{d}$ are model parameters. ReLU \citep{nair2010rectified} is the activation function, and $\oplus$ refers to the vector concatenation. 

To reduce the error propagation issue, following \citet{DBLP:conf/acl/ZhangZWLZ20}, we use the arc probability matrix obtained from GOPar as the adjacency matrix $A$, which may provide richer  syntactic structures. For $\mathbf{e}^{(i,j)}$, we use %
the 1-best label of the 1-best head word $w_j$ %
of $w_j$.

We then feed the outputs of the GCN sub-layer to the feed-forward (FF) sub-layer that contains two linear transformations with a ReLU activation function in between, as shown below:
\begin{equation}
    \mbox{FF}(\mathbf{h})=\mbox{ReLU}(W_1 \mathbf{h} + \mathbf{b_1})W_2 + \mathbf{b_2}
\end{equation}

\subsection{Representation Fusion} 

In order to balance the contribution of the syntax-aware representations from DepGCN ($\mathbf{h}_i^{syn}$) and the representations from the basic Transformer encoder ($\mathbf{h}_i^{basic}$), we use their interpolation (i.e., weighted-sum) as the final representations, which are ultimately fed into the Transformer decoder:
\begin{equation}
    \mathbf{h}_i^{final}= \beta {\mathbf{h}_i^{basic}} + (1-\beta)\mathbf{h}_i^{syn}
    \label{eq:agg}
\end{equation}
where $\beta\in(0,1)$ is a hyper-parameter called the fusion factor, and $\mathbf{h}_i^{final}$ represents the final output vector of the Transformer encoder for the $i$-th token. 
As depicted in Figure \ref{fig:model}, 
this operation is analogous to the residual connection.

\section{Experiments on English GEC}
\begin{table}[tp!]
\scalebox{0.7}{
\begin{tabular}{lccc}
\toprule
\textbf{Dataset} & \textbf{\#Sentences} & \textbf{\%Error} & \textbf{Usage}    \\ \hline
\textbf{CLang8}            & 2,372,119            & 57.8             & Pre-training         \\ 
\textbf{FCE}              & 34,490               & 62.6             & Fine-tuning I     \\
\textbf{NUCLE}            & 57,151               & 38.2             & Fine-tuning I     \\
\textbf{W\&I+LOCNESS}     & 34,308               & 66.3             & Fine-tuning I\&II \\
\hline
\textbf{BEA-19-\textit{Dev}}              & 4,384               & 65.2             & Validation     \\ 
\textbf{CoNLL-14-\textit{Test}}              & 1,312               & 72.3             & Testing     \\
\textbf{BEA-19-\textit{Test}}             & 4,477               & -             & Testing     \\ 
\bottomrule
\end{tabular}
}
\caption{Statistics of English GEC datasets. \textbf{\#Sentences} denotes the number of sentences.\textbf{\%Error} refers to the proportion of erroneous sentences.}
\label{tab:dataset}
\end{table}

\begin{table*}[tp!]
\centering
\scalebox{0.7}{
\begin{tabular}{clcccccccc}
\toprule
                                  &  &\textbf{Extra}  & \multicolumn{1}{c}{\textbf{Transformer}}         & \multicolumn{3}{c}{\textbf{CoNLL-14-\textit{test}}}      & \multicolumn{3}{c}{\textbf{BEA-19-\textit{test}}}     \\
                                  & \textbf{System}& \textbf{Data Size} & \textbf{Layer, Hidden, FFN}  & \textbf{P}    & \textbf{R} & $\mbox{\textbf{F}}_{0.5}$ & \textbf{P} & \textbf{R} &$\mbox{\textbf{F}}_{0.5}$ \\ \hline
\multirow{10}{*}{\textbf{w/o PLM}} & \multicolumn{9}{l}{\textbf{w/o syntax}}                                                                  \\
                                  & \citet{kiyono2019empirical}$^{\circ}$ &   70M &   12+12,1024,4096         & 67.9          & 44.1       & 61.3          & 65.5       & 59.4       & 64.2          \\
                                  & \citet{lichtarge-etal-2020-data}$^{\vartriangle\blacktriangle}$&   340M &   12+12,1024,4096          & 69.4          & 43.9       & 62.1          & 67.6       & 62.5       & 66.5          \\
                                  & \citet{stahlberg2021synthetic}$^{\vartriangle\blacktriangle\square}$&   540M &   12+12,1024,4096         & 72.8          & 49.5       & \textbf{66.6}          & 72.1       & 64.4       & \textbf{70.4}          \\
                                  & \textbf{Our Baseline}$^{\heartsuit}$&   2.4M &   6+6,512,2048  & 66.9          & 	40.3      & 	59.1          & 66.8      &	55.5       &	64.2         \\ 
                                  \cline{2-10} 
                                  & \multicolumn{7}{l}{\textbf{w/ syntax}}                                                                   \\
                                  & \citet{wan2021syntax}$^{\blacklozenge}$&   10M &   6+6,512,2048                & 74.4          & 39.5       & 63.2          & 74.5       & 48.6       & 67.3          \\
                                & \citet{li2022syntax}$^{\spadesuit}$&   30M &   12+12,1024,4096               & 66.7 &  38.3  & 58.1          & -       & -       & -          \\
                                  & \textbf{SynGEC}$^{\heartsuit}$&   2.4M &   6+6,512,2048  & 70.0          & 46.2       & \textbf{63.5}          & 70.9       & 59.9       & \textbf{68.4}          \\
                                   & \hspace{0.3cm}\textbf{GOPar$\rightarrow$Off-the-shelf Parser}&   2.4M &   6+6,512,2048  & 68.2          & 40.9       & 60.2           & 67.3       & 55.4       & 64.5          \\\hline \hline
\multirow{12}{*}{\textbf{w/ PLM}}  & \multicolumn{9}{l}{\textbf{w/o syntax}}                                                                  \\
                                    & \citet{kaneko2020encoder}$^{\circ}$&   70M &   12+12,1024,4096            & 69.2          & 45.6       & 62.6          & 67.1       & 60.1       & 65.6          \\
                                  & \citet{katsumata2020stronger}&   - &   12+12,1024,4096              & 69.3          & 45.0       & 62.6          & 68.3       & 57.1       & 65.6          \\
                                  & \citet{omelianchuk2020gector}$^{\lozenge}$&   9M &   12+0,768,3072            & 77.5          & 40.1       & 65.3          & 79.2       & 53.9       & 72.4          \\
                                   & \citet{rothe2021recipe}$^{\heartsuit}$&   2.4M &   12+12,1024,4096            & -             & -          & 66.1         & -          & -          & 72.1          \\
                                  & \citet{DBLP:conf/acl/SunGWW20}&   300M &   12+2,1024,4096            & 71.0             & 52.8          & 66.4         & -          & -          & \textbf{72.9}          \\
                                    & \textbf{Our Baseline}$^{\heartsuit}$&   2.4M &   12+12,1024,4096 &  73.6          & 	48.6      & 	\textbf{66.7}         & 74.0       & 64.9       & 72.0          \\ 
                                  \cline{2-10} 
                                  & \multicolumn{9}{l}{\textbf{w/ syntax}}                                                                   \\ 
                                & \citet{li2022syntax}$^{\spadesuit}$&   30M &   12+12,1024,4096               & 68.1 &  44.1  & 61.4          & -       & -       & -          \\
                                  & \textbf{SynGEC}$^{\heartsuit}$&   2.4M  &   12+12,1024,4096 & 74.7 & 49.0      & \textbf{67.6}          & 75.1      & 65.5      & \textbf{72.9}          \\ 
                                   & \hspace{0.3cm}\textbf{GOPar$\rightarrow$Off-the-shelf Parser}&   2.4M &   12+12,1024,4096 & 74.1          & 48.3       & 67.0          & 74.6       & 64.1       & 72.3          \\
\bottomrule
\end{tabular}
}
\caption{\textbf{Single-model} results on English GEC test-sets. Our results are averaged over three runs with different random seeds. \textbf{Layer}, \textbf{Hidden} and \textbf{FFN} denote the depth, hidden size and feed-forward network size of Transformer. ``\textbf{w/ PLM}'' means using pre-trained language models. ``\textbf{w/ syntax}'' means using syntactic knowledge. Besides the public human-annotated training data, current GEC systems variously use private and/or artificial data, including: $^{\circ}$artificial Gigaword (70M sentences), $^{\vartriangle}$Wikipedia revision histories (170M), $^{\blacktriangle}$artificial Wikipedia (170M), $^{\square}$artificial Colossal Clean Crawled Corpus (200M), $^{\lozenge}$artificial one-billion-word (9M), $^{\blacklozenge}$artificial one-billion-word (10M), $^{\spadesuit}$artificial one-billion-word (30M) , $^{\heartsuit}$cleaned version of Lang8 (2.4M).}

\label{tab:main:results}
\end{table*}
\subsection{Settings}
\label{sec:english:exp}
\textbf{Datasets and evaluation.} We first pre-train our model on the cleaned version of the Lang8 dataset (CLang8) \footnote{CLang8 can be downloaded from \url{https://github.com/google-research-datasets/clang8}} released by \citet{rothe2021recipe}. Then, we use the FCE dataset \citep{yannakoudakis2011new}, the NUCLE dataset \citep{dahlmeier2013building} and the W\&I+LOCNESS train-set \citep{bryant2019bea} for model fine-tuning following previous studies. Like \citet{omelianchuk2020gector}, we decompose the fine-tuning procedure into two stages: 1) fine-tuning on FCE+NUCLE+W\&I+LOCNESS; 2) further fine-tuning only on the small-scale but high-quality W\&I+LOCNESS. 

For evaluation, we report average P/R/F$_{0.5}$ results over three runs with different random seeds on the CoNLL-14 test set\footnote{We use the \textit{official-2014.combined.m2} (no-alt) version of CoNLL-14, which is adopted by most existing works.}
\citep{ng2014conll} evaluated by M2Scorer \citep{dahlmeier2012better} and BEA-19 test set \citep{bryant2019bea} evaluated by ERRANT \citep{bryant2017automatic}. The BEA-19 dev set serves as validation data during the whole training. The statistics of above datasets are shown in Table \ref{tab:dataset}. 

Besides, we also experiment on the small-scale JFLEG test-set \cite{napoles2017jfleg} and list the results in Appendix \ref{sec:appendix:jfleg}.

\textbf{GEC model details.}
We adopt \texttt{Fairseq}\footnote{\url{https://github.com/pytorch/fairseq}} \citep{ott2019fairseq} to build our Transformer baseline and DepGCN-based model. 
For the DepGCN-based model, we empirically stack $N_2 = 3$ DepGCN blocks and set the fusion factor $\beta = 0.5$ in Equation \ref{eq:agg}. 
We apply BPE \citep{sennrich-etal-2016-neural} to generate a 32K shared subword vocabulary. 
We apply the Dropout-Src mechanism \cite{junczys2018approaching} to source-side word embeddings  
to alleviate over-fitting. 
More model details are discussed in Appendix \ref{sec:appendix:hp}.

\textbf{GOPar details.} 
The training data for GOPar is generated from the CLang8 dataset \citep{rothe2021recipe} using the procedure described in Section~\ref{sec:parsing:data}. 
For all English experiments, GOPar operates at the word level. 
In contrast, our GEC models perform on the subword level. %
To fill this gap, we transform word-level syntax trees into subword-level ones by adding arcs. For example, if $w_i$ is the head word of $w_j$, we will 
add arcs from all subwords of $w_i$ to all subwords of $w_j$. All added arcs copy the arc probability of $w_i \rightarrow w_j$. We will explore more sophisticated ways to handle this mismatch  issue.

For both the off-the-shelf parser and GOPar, we use the biaffine parsing approach \citep{DBLP:conf/iclr/DozatM17}.
We directly adopt the implementation of  \texttt{SuPar}\footnote{\url{https://github.com/yzhangcs/parser}} \citep{DBLP:conf/acl/ZhangLZ20} and follow their default hyper-parameter settings. 
After comparing several popular PLMs, 
we choose to use 
ELECTRA \citep{DBLP:conf/iclr/ClarkLLM20} to provide the contextual token representations for parsers. 
To obtain word-level representations, we aggregate the subword-level representations from ELECTRA into word-level ones via average pooling. 
The off-the-shelf parser is trained on PTB \cite{marcinkiewicz1994building}. 
Please kindly notice that all parsers are always enhanced with ELECTRA, even when the GEC model does not use PLM.

\textbf{Incorporating BART.} 
To explore whether the syntactic knowledge is still useful after introducing powerful PLMs, we use BART \citep{lewis2020bart} to initialize the Transformer backbone of our models.
It is noteworthy that we adopt a two-stage training procedure to keep the training stable. Firstly, we fine-tune the BART-initialized Transformer backbone until it converges. Secondly, we add an auxiliary DepGCN module into the converged Transformer backbone and only tune the DepGCN parameters on the same training data. The intuition behind this procedure is that the BART part has been extensively pre-trained whereas the DepGCN part is just randomly initialized. In our preliminary experiments, we frequently encountered training collapse when training two parts simultaneously, and observed that the scale of the gradients of the two parts vary substantially.

\label{sec:res}
\subsection{Main Results}
The main results are listed in Table \ref{tab:main:results}. 
In the top group of results without PLMs, SynGEC achieves 63.5/68.4 $\mbox{F}_{0.5}$ scores on CoNLL-14 and BEA-19 test-sets, respectively, outperforming all other systems utilizing syntax.
The performance of SynGEC is only lower than \citet{stahlberg2021synthetic} on both test-sets, probably because they use an extra huge synthetic corpus with 540M sentence-pairs. The incorporation of syntactic information provided by GOPar leads to 4.4/4.2 $\mbox{F}_{0.5}$ improvements 
over our baseline, which demonstrates that the tailored syntactic knowledge from GOPar is quite helpful for GEC and our DepGCN-based GEC model can effectively capture it. 

In the bottom group of results using PLMs, our SynGEC approach augmented with BART achieves 67.6/72.9 $\mbox{F}_{0.5}$ scores, which are comparable or even better than other cutting-edge PLM-enhanced models under similar sizes.
After removing syntax, the $\mbox{F}_{0.5}$ scores decline by 0.9 on both datasets, which reveals that the contribution from adaptive syntax and PLMs does not fully overlap. It is worth noting that \citet{rothe2021recipe} also build another much larger GEC model based on the T5-11B \citep{raffel2020exploring} and achieve 68.9/75.9  $\mbox{F}_{0.5}$ scores. For a fair comparison, we do not list this result in Table \ref{tab:main:results} as this model is about 24$\times$ larger than ours.

\subsection{Analysis and Discussion}
\label{ref:sec:analysis}

\label{sec:ana}
\textbf{Effectiveness of GOPar.} In Table \ref{tab:main:results}, we present the results of using an off-the-shelf parser\footnote{We use \href{https://github.com/yzhangcs/parser/releases/download/v1.1.0/ptb.biaffine.dep.roberta.zip}{\textit{biaffine-dep-roberta-en}} model provided by SuPar.} 
to provide syntactic knowledge (GOPar $\rightarrow$ Off-the-shelf Parser). After changing the parser, the impact of syntax becomes marginal under all settings. This observation implies that the performance gains contributed from syntax are highly contingent on the quality of parses. We look further into the parses and find that GOPar is more robust when facing grammatical errors and can further identify such errors, while the off-the-shelf parser is vulnerable and tends to provide incorrect parses. So we can draw a conclusion that the task adaptation of parsers is essential when applying syntax to the GEC task.

\begin{table}[tp!]
\centering
\scalebox{0.75}{
\begin{tabular}{lcc}
\toprule
& \textbf{BEA-19-\textit{dev}}& \textbf{CoNLL-14-\textit{test}} \\ & \textbf{P}/\textbf{R}/\textbf{$\mbox{\textbf{F}}_{0.5}$} & \textbf{P}/\textbf{R}/\textbf{$\mbox{\textbf{F}}_{0.5}$} \\\hline 
\multicolumn{3}{c}{\textbf{w/o PLM}} 
 \\
\hline 
\textbf{SynGEC}&  60.84/\textbf{39.76}/\textbf{55.01} & 70.03/\textbf{46.17}/\textbf{63.47}         \\
\hspace{0.3cm}\textbf{w/o GED Labels}   & 59.47/38.03/53.44 &  69.79/43.71/62.35        \\
\hspace{0.3cm}\textbf{w/o Syntax Labels}   & 59.31/39.02/53.72  &   69.31/44.65/62.42    \\
\hspace{0.3cm}\textbf{w/o All Labels}  & \textbf{61.62}/34.21/53.11  &       \textbf{71.14}/40.02/61.57
\\ 
\textbf{Baseline}  & 57.99/35.77/51.58  &   66.94/40.25/59.10
\\\hline\hline
\multicolumn{3}{c}{\textbf{w/ PLM}}  \\
\hline \textbf{SynGEC}    & 64.51/\textbf{45.73}/\textbf{59.62}  &     74.67/\textbf{48.98}/\textbf{67.58}   \\
\hspace{0.3cm}\textbf{w/o GED Labels}       &  63.59/45.23/58.82 &   74.03/48.65/67.04    \\
\hspace{0.3cm}\textbf{w/o Syntax Labels}    & 64.20/45.51/59.33 &    73.95/48.87/67.05      \\
\hspace{0.3cm}\textbf{w/o All Labels}      &  \textbf{64.90}/42.57/58.74 &   \textbf{75.11}/46.72/66.97  \\
\textbf{Baseline}   &   63.09/44.80/58.32 &  73.62/48.58/66.74
\\
\bottomrule
\end{tabular}
}
\caption{Effect of different syntactic information. Since BEA-19-\textit{test} need online submission, we instead report results on BEA-19-\textit{dev}. %
}
\label{tab:scheme}
\end{table}

\begin{figure*}[ht!]
\centering
\includegraphics[scale=0.55]{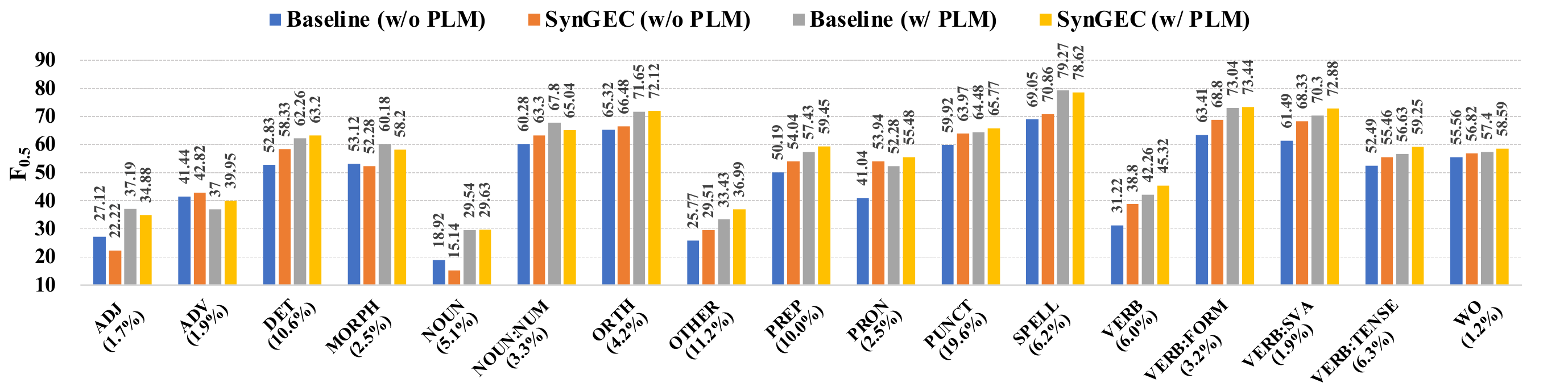}
\caption{Performance on a selection of error types from ERRANT \citep{bryant2017automatic} on BEA-19-\textit{dev}. Numbers in parentheses represent the percentages of error types. We exclude error types that account for less than 1\%.}
\label{fig:error:type}
\end{figure*}

\textbf{Decomposition of syntactic information.} To gain more insights on how adaptive syntactic information works, we decompose it into three parts: 1) the arc information, which means only using the topological structure of the syntax tree; 2) the GED label information, which refers to the special labels ``S'', ``R'' and ``M'' for marking erroneous tokens; and 3) the syntax label information, such as ``subj'' and ``iobj'' for different syntactic relations. We conduct an ablation study to explore the effect of each kind of information for GEC, as shown in Table \ref{tab:scheme}. Concretely, for ``w/o GED Labels'', we force the parser to skip GED labels and select the syntax label with the highest probability when predicting. For ``w/o Syntax Labels'', we replace all syntax labels with ``O'' in the results. For ``w/o All Labels'', we do not feed the label embeddings into the DepGCN module and use the dependency distance information to re-scale the self-attention weights in the Transformer encoder.

There are several observations. First, removing GED labels or syntax labels reduces the performance of SynGEC to a similar extent, which indicates that they are equally important to GEC. 
Second, when we only use the arc information, i.e., removing all labels, the recall drops sharply while the precision increases notably compared with the baseline. We speculate that the contribution from arc information is mainly on preventing GEC models from being misled by inappropriate context. 
Third, the full SynGEC approach utilizing all three kinds of information achieves the best performance, which implies that they have intrinsic complementary strengths.

\begin{table}[tp!]
\centering
\scalebox{0.75}{
\begin{tabular}{lcc}
\toprule
& \textbf{BEA-19-\textit{dev}}& \textbf{CoNLL-14-\textit{test}} \\ & \textbf{P}/\textbf{R}/\textbf{$\mbox{\textbf{F}}_{0.5}$} & \textbf{P}/\textbf{R}/\textbf{$\mbox{\textbf{F}}_{0.5}$} \\\hline 
\multicolumn{3}{c}{\textbf{w/o PLM}} 
 \\
\hline 
\textbf{Baseline}  & 57.99/35.77/51.58  &   66.94/40.25/59.10 \\
\textbf{Self-training}  & 58.85/36.30/52.35  &   67.98/40.93/60.04 \\
\textbf{SynGEC}&  \textbf{60.84}/\textbf{39.76}/\textbf{55.01} & \textbf{70.03}/\textbf{46.17}/\textbf{63.47}  \\
\hline\hline
\multicolumn{3}{c}{\textbf{w/ PLM}}  \\
\hline 
\textbf{Baseline}   &   63.09/44.80/58.32 &  73.62/48.58/66.74 \\
\textbf{Self-training}  &   63.49/44.37/58.45 &  74.15/48.32/66.99 \\  
\textbf{SynGEC}    & \textbf{64.51}/\textbf{45.73}/\textbf{59.62}  &     \textbf{74.67}/\textbf{48.98}/\textbf{67.58}   \\
\bottomrule
\end{tabular}
}
\caption{Comparison with the self-training method.
}
\label{tab:self-training}
\end{table}

\textbf{Influence of self-training.} Despite the effectiveness of GOPar compared with off-the-shelf parsers trained on small-scale manually-annotated treebanks of grammatical sentences, it is still not clear whether---or to what extent---the improvement comes from our GEC-oriented adaption of the parser. It is also possible that some or most improvement is due to the larger training set and domain adaptation via self-training \cite{DBLP:conf/naacl/McCloskyCJ06}. Self-training is a classical semi-supervised learning method that enhances models with large-scale pseudo-labeled in-domain data. To study this, we directly utilize the pseudo-labeled trees of target-side sentences in GEC data to train a parser (\textbf{Self-training} in Table \ref{tab:self-training}). We observe that SynGEC significantly outperforms only using self-training without the step of tree projection, which demonstrates that the effectiveness of GOPar mainly stems from the GEC-oriented adaptation.

\textbf{Error type performance.} Figure \ref{fig:error:type} shows more fine-grained evaluation results on different error types on BEA-19-\textit{dev}.
The results support that syntactic information from GOPar is beneficial for most error types. Specifically, syntactic knowledge significantly improves the GEC model's ability to correct context-sensitive errors, such as DET, PREP, PUCNT, VERB:SVA, and VERB:TENSE. Correcting such errors requires long-distance information, which syntax can effectively provide. The syntax also helps solve word-ordering (WO) errors, which need sentence structure information to correct. Besides, the performance on PRON, OTHER, VERB is also substantially improved. Meanwhile, we note that a small subset of types is negatively affected, like ADJ, MORPH, SPELL, and NOUN. After more careful observation, we find that their corrections mainly depend on local information, where syntactic knowledge may not help much or even introduce noises.

\section{Experiments on Chinese GEC}
\label{sec:chinese:exp}

\textbf{Datasets and evaluation.} 
For Chinese, we report P/R/$\mbox{F}_{0.5}$ values on NLPCC-18-\textit{test} \citep{zhao2018overview} and MuCGEC-\emph{test} \cite{zhang2022mucgec} using their official evaluation tools. MuCGEC-\emph{dev} is used for hyper-parameter tuning and checkpoint selection. For \emph{training} data, we use the Chinese Lang8 dataset \citep{zhao2018overview} and HSK dataset \citep{zhang2009hsk}. The statistics of above-mentioned datasets are shown in Table \ref{tab:chinese:dataset}.

\begin{table}[tp!]\
\centering
\scalebox{0.8}{
\begin{tabular}{lccc}
\toprule
\textbf{Dataset} & \textbf{\#Sentences} & \textbf{\%Error} & \textbf{Usage}    \\ \hline
\textbf{Lang8}            &   1,220,906          & 89.5             & Training         \\ 
\textbf{HSK}            & 15,6870            & 60.8             & Training         \\ 
\hline
\textbf{MuCGEC-\textit{dev}}            & 1,125            & 95.1             & Validation         \\ 
\textbf{MuCGEC-\textit{test}}            & 5,938            & 92.2             & Testing         \\ 
\textbf{NLPCC-18-\textit{test}}            & 2,000            & 99.2             & Testing         \\ 
\bottomrule
\end{tabular}
}
\caption{Statistics of Chinese GEC datasets.}
\label{tab:chinese:dataset}
\end{table}
\textbf{Char-based GOPar.} 
Current Chinese GEC models usually treat the input sentence as a character sequence and do not perform word segmentation \cite{zhao2020maskgec}. In contrast, dependency parsers typically treat the input sentence as a word sequence. 
To handle this mismatch, we follow \citet{DBLP:journals/tacl/YanQH20}  and build a char-based GOPar. The basic idea is to convert a word-based tree into a char-based one by letting each character depends on its right-hand one  inside multi-character words.

\textbf{Use of BART.} 
We employ the recently proposed Chinese BART \citep{shao2021cpt}, which is originally implemented with the HuggingFace Transformers toolkit\footnote{\url{https://github.com/huggingface/transformers}} \citep{wolf-etal-2020-transformers}. 
We manage to wrap their code and use it on our Fairseq implementation.
Specifically, we find  many common characters are missing in its vocabulary. Therefore, we add 3,866 Chinese characters and punctuation marks from Chinese Gigaword and Wikipedia corpora, leading to a substantial performance boost according to our preliminary experiments. The embeddings of these newly added tokens are randomly initialized and trained on GEC data.

\begin{table}[tp!]
\centering
\scalebox{0.7}{
\begin{tabular}{lccccc}
\hline
                          & \textbf{PLM} & \textbf{Syntax} & \textbf{P}     & \textbf{R}     & $\mbox{\textbf{F}}_{0.5}$              \\ \hline 
\multicolumn{6}{c}{\textbf{NLPCC-18-\textit{test}}}  \\ \hline
\textbf{\citet{zhang2022mucgec}}                      &  \cmark   &   \xmark     & 42.88 & 30.19 & 39.55          \\ \hline
\textbf{Baseline} &    \xmark  &     \xmark   & 40.10 & 26.34 & 36.31           \\
\textbf{SynGEC}                           &   \xmark   &  \cmark      & 41.44 & 28.28 & \textbf{37.91} \\ \hline
\textbf{Baseline}                          &   \cmark  &   \xmark      & 49.07 & 32.80 & 44.64      \\
\textbf{SynGEC}                           & \cmark    &  \cmark      & 49.96 & 33.04 & \textbf{45.32} \\ \hline\hline
                          \multicolumn{6}{c}{\textbf{MuCGEC-\textit{test}}}  \\ \hline
\textbf{\citet{zhang2022mucgec}}                      &  \cmark   &   \xmark     & 43.81 & 28.56 & 39.58          \\ \hline
\textbf{Baseline} &    \xmark  &     \xmark    & 43.79 & 25.93 & 38.49          \\
\textbf{SynGEC}                           &   \xmark   &  \cmark      & 46.88 & 27.68 & \textbf{40.58} \\ \hline
\textbf{Baseline}                          &   \cmark  &   \xmark      & 54.21 & 28.51 & 45.93          \\
\textbf{SynGEC}                           & \cmark    &  \cmark      & 54.69 & 29.10 & \textbf{46.51} \\ \hline
\end{tabular}
}
\caption{Single-model results on Chinese datasets. %
}
\label{tab:chinese:result}
\end{table}

\textbf{Results} are presented in Table \ref{tab:chinese:result}. When not using BART, our SynGEC outperforms the Transformer baseline by  1.60/2.09 $\mbox{F}_{0.5}$ score on NLPCC-18-\textit{test} and MuCGEC-\textit{test}, respectively. When using BART, our baseline already outperforms the previous SOTA system \citep{zhang2022mucgec}, thanks to the engineering efforts mentioned in the previous paragraph. Again, SynGEC further improves the $\mbox{F}_{0.5}$ score by 0.68/0.58. These results indicate that our proposed SynGEC approach can be effective for different languages. Given that our SynGEC approach is actually language-independent, we plan to test it in more languages in the future.

\section{Related Works}
\label{relate}

\textbf{Grammatical error correction.} 
Recent work mainly formulates GEC as a monolingual translation task and handle it with burgeoning encoder-decoder-based MT models \citep{yuan2016grammatical, junczys2018approaching}, among which Transformer \citep{vaswani2017attention} has become a dominant paradigm. With the help of synthetic training data  \citep{lichtarge2019corpora, DBLP:conf/emnlp/YasunagaLL21} and large PLMs \citep{kaneko2020encoder, katsumata2020stronger}, Transformer-based GEC models have achieved SOTA performance on various benchmark datasets \citep{rothe2021recipe,stahlberg2021synthetic}.

Meanwhile, the sequence-to-edit (Seq2Edit) approach emerges as a competitive alternative, which predicts a sequence of edit operations to achieve correction \citep{gu2019levenshtein,awasthi2019parallel,omelianchuk2020gector}. 
Although this work adopts the Transformer-based GEC models as the baseline, our SynGEC approach can also be applied to Seq2Edit models straightforwardly, which we leave to future work.

\textbf{Parsing ungrammatical sentences.} Despite the success of syntactic parsing on clean sentences \citep{DBLP:conf/iclr/DozatM17, DBLP:conf/acl/ZhangLZ20}, parsing noisy sentences is still under-explored, including but not limited to learner texts \citep{DBLP:conf/lrec/Foster04,hashemi2016evaluation}, speech disfluencies \cite{honnibal2014joint}, and historical texts \cite{pettersson2012parsing}. This work focuses on parsing ungrammatical texts. Previous studies mainly tackle this problem by annotating small-scale trees for ungrammatical sentences and re-training a parser on them \citep{dickinson2009dependency,petrov2012overview,cahill2015parsing,DBLP:conf/acl/BerzakKSWLMGK16}. 
Instead, we propose to train a tailored parser on automatically generated syntax trees from parallel GEC data, which avoids the laborious manual annotation. This idea has been mentioned as future work in \citet{wagner2012detecting}.

\textbf{Syntax-enhanced GEC.} 
Many previous work has demonstrated the effectiveness of utilizing syntactic information for various NLP tasks,  such as machine translation \citep{bastings-etal-2017-graph,DBLP:conf/naacl/ZhangLFZ19}, opinion role labeling \citep{DBLP:conf/acl/ZhangZWLZ20}, and semantic role labeling \citep{xia2019syntax,sachan2021syntax}.
Meanwhile, we have found two recent works on syntax-enhanced GEC \citep{wan2021syntax,li2022syntax}.  Both works directly produce the dependency tree of the input sentence using an off-the-shelf parser, without tailoring parsers for ungrammatical sentences. 
They both use graph attention networks (GAT) for tree encoding \citep{DBLP:conf/iclr/VelickovicCCRLB18}. 
Besides the dependency tree, \citet{li2022syntax} exploits the constituent tree of the input sentence as well.  

Compared with the above two works, the major contribution of our work is directly dealing with the severe performance drop issue via our tailored GOPar. 
We adopt GCN for tree encoding because our preliminary experiments show that it achieves similar performance but is faster. 
Moreover, our baseline GEC models achieve much higher performance than theirs, as shown in Table \ref{tab:main:results}.

\section{Conclusions}

This paper presents a SynGEC approach that incorporates adapted dependency syntax into GEC models. The key idea is adjusting vanilla parsers to accommodate ungrammatical sentences. We first extend the standard syntax representation scheme to use a unified tree structure to encode both grammatical errors and syntactic structure. Then we obtain high-quality parse trees of ungrammatical sentences by projecting target-side trees into source-side ones in parallel GEC training data, which are ultimately used for training a tailored parser named GOPar.
We employ GCN to encode syntax produced by GOPar. Experiments on mainstream datasets in two languages show that SynGEC is effective and achieves SOTA results.

\section*{Limitations}
First, off-the-shelf parsers may still produce noisy parse trees for the target-side correct sentences, which could further lead to noise in our projected trees for the source-side incorrect sentences. Second, we have only employed three coarse-grained labels to distinguish grammatical errors in our syntax representation scheme, while fine-grained categories may further benefit GEC \citep{DBLP:conf/emnlp/YuanTD021}. Both limitations may be mitigated by integrating ideas and resources achieved by previous work on manually annotating syntactic trees for ungrammatical sentences \citep{dickinson2009dependency,DBLP:conf/acl/BerzakKSWLMGK16}. Besides, all of our efforts focus on integrating source-side syntactic information, while there is also some work trying to incorporate target-side syntax and get positive results \citep{aharoni2017towards,wang2018tree}. We will further study how to appropriately utilize such target-side syntax in our future work. 

\section*{Acknowledgements}
We want to thank all the anonymous reviewers for their valuable comments. We also thank Yu Zhang, Houquan Zhou for their great help and insightful suggestions when polishing this paper. This work was partially supported by the National Natural Science Foundation of China (Grant No.62176173 and No.61876116) and by Alibaba Group through Alibaba Innovative  Research Program. This work was also partially supported by Projected Funded by the Priority Academic Program Development of Jiangsu Higher Education Institutions.

\bibliography{anthology,custom}

\begin{thebibliography}{77}
\expandafter\ifx\csname natexlab\endcsname\relax\def\natexlab#1{#1}\fi

\bibitem[{Aharoni and Goldberg(2017)}]{aharoni2017towards}
Roee Aharoni and Yoav Goldberg. 2017.
\newblock Towards string-to-tree neural machine translation.
\newblock In \emph{Proceedings of ACL (Short Papers)}, pages 132--140.

\bibitem[{Awasthi et~al.(2019)Awasthi, Sarawagi, Goyal, Ghosh, and
  Piratla}]{awasthi2019parallel}
Abhijeet Awasthi, Sunita Sarawagi, Rasna Goyal, Sabyasachi Ghosh, and Vihari
  Piratla. 2019.
\newblock Parallel iterative edit models for local sequence transduction.
\newblock In \emph{Proceedings of EMNLP-IJCNLP}, pages 4260--4270.

\bibitem[{Ba et~al.(2016)Ba, Kiros, and Hinton}]{DBLP:journals/corr/BaKH16}
Lei~Jimmy Ba, Jamie~Ryan Kiros, and Geoffrey~E. Hinton. 2016.
\newblock Layer normalization.
\newblock \emph{arXiv preprint arXiv:1607.06450}.

\bibitem[{Bastings et~al.(2017)Bastings, Titov, Aziz, Marcheggiani, and
  Sima{'}an}]{bastings-etal-2017-graph}
Jasmijn Bastings, Ivan Titov, Wilker Aziz, Diego Marcheggiani, and Khalil
  Sima{'}an. 2017.
\newblock Graph convolutional encoders for syntax-aware neural machine
  translation.
\newblock In \emph{Proceedings of EMNLP}, pages 1957--1967.

\bibitem[{Bell et~al.(2019)Bell, Yannakoudakis, and
  Rei}]{bell-etal-2019-context}
Samuel Bell, Helen Yannakoudakis, and Marek Rei. 2019.
\newblock Context is key: Grammatical error detection with contextual word
  representations.
\newblock In \emph{Proceedings of {BEA}@ACL}, pages 103--115.

\bibitem[{Berzak et~al.(2016)Berzak, Kenney, Spadine, Wang, Lam, Mori, Garza,
  and Katz}]{DBLP:conf/acl/BerzakKSWLMGK16}
Yevgeni Berzak, Jessica Kenney, Carolyn Spadine, Jing~Xian Wang, Lucia Lam,
  Keiko~Sophie Mori, Sebastian Garza, and Boris Katz. 2016.
\newblock Universal dependencies for learner {English}.
\newblock In \emph{Proceedings of ACL}, pages 737--746.

\bibitem[{Bryant et~al.(2019)Bryant, Felice, Andersen, and
  Briscoe}]{bryant2019bea}
Christopher Bryant, Mariano Felice, {\O}istein~E Andersen, and Ted Briscoe.
  2019.
\newblock The {BEA}-2019 shared task on grammatical error correction.
\newblock In \emph{Proceedings of {BEA}@ACL}, pages 52--75.

\bibitem[{Bryant et~al.(2017)Bryant, Felice, and Briscoe}]{bryant2017automatic}
Christopher Bryant, Mariano Felice, and Ted Briscoe. 2017.
\newblock Automatic annotation and evaluation of error types for grammatical
  error correction.
\newblock In \emph{Proceedings of ACL}, pages 793--805.

\bibitem[{Cahill(2015)}]{cahill2015parsing}
Aoife Cahill. 2015.
\newblock Parsing learner text: {T}o shoehorn or not to shoehorn.
\newblock In \emph{Proceedings of Linguistic Annotation Workshop}, pages
  144--147.

\bibitem[{Clark et~al.(2020)Clark, Luong, Le, and
  Manning}]{DBLP:conf/iclr/ClarkLLM20}
Kevin Clark, Minh{-}Thang Luong, Quoc~V. Le, and Christopher~D. Manning. 2020.
\newblock {ELECTRA:} pre-training text encoders as discriminators rather than
  generators.
\newblock In \emph{Proceedings of ICLR}.

\bibitem[{Cui et~al.(2020)Cui, Che, Liu, Qin, Wang, and Hu}]{cui2020revisiting}
Yiming Cui, Wanxiang Che, Ting Liu, Bing Qin, Shijin Wang, and Guoping Hu.
  2020.
\newblock Revisiting pre-trained models for {Chinese} natural language
  processing.
\newblock In \emph{Proceedings of EMNLP: findings}, pages 657--668.

\bibitem[{Dahlmeier and Ng(2012)}]{dahlmeier2012better}
Daniel Dahlmeier and Hwee~Tou Ng. 2012.
\newblock Better evaluation for grammatical error correction.
\newblock In \emph{Proceedings of NAACL-HLT}, pages 568--572.

\bibitem[{Dahlmeier et~al.(2013)Dahlmeier, Ng, and Wu}]{dahlmeier2013building}
Daniel Dahlmeier, Hwee~Tou Ng, and Siew~Mei Wu. 2013.
\newblock Building a large annotated corpus of learner {English}: The nus
  corpus of learner {English}.
\newblock In \emph{Proceedings of {BEA}@NAACL-HLT}, pages 22--31.

\bibitem[{Dickinson and Ragheb(2009)}]{dickinson2009dependency}
Markus Dickinson and Marwa Ragheb. 2009.
\newblock Dependency annotation for learner corpora.
\newblock In \emph{Proceedings of International Workshop on Treebanks and
  Linguistic Theories}, page~59.

\bibitem[{Dozat and Manning(2017)}]{DBLP:conf/iclr/DozatM17}
Timothy Dozat and Christopher~D. Manning. 2017.
\newblock Deep biaffine attention for neural dependency parsing.
\newblock In \emph{Proceedings of ICLR}.

\bibitem[{Foster(2004)}]{DBLP:conf/lrec/Foster04}
Jennifer Foster. 2004.
\newblock Parsing ungrammatical input: an evaluation procedure.
\newblock In \emph{Proceedings of LREC}.

\bibitem[{Foster et~al.(2008)Foster, Wagner, and
  Van~Genabith}]{foster2008adapting}
Jennifer Foster, Joachim Wagner, and Josef Van~Genabith. 2008.
\newblock Adapting a wsj-trained parser to grammatically noisy text.
\newblock In \emph{Proceedings of ACL (short)}, pages 221--224.

\bibitem[{Grundkiewicz et~al.(2020)Grundkiewicz, Bryant, and
  Felice}]{grundkiewicz2020crash}
Roman Grundkiewicz, Christopher Bryant, and Mariano Felice. 2020.
\newblock A crash course in automatic grammatical error correction.
\newblock In \emph{Proceedings of COLING: Tutorial Abstracts}, pages 33--38.

\bibitem[{Gu et~al.(2019)Gu, Wang, and Junbo}]{gu2019levenshtein}
Jiatao Gu, Changhan Wang, and Jake~Zhao Junbo. 2019.
\newblock Levenshtein transformer.
\newblock In \emph{Proceedings of NIPS}, pages 11181--11191.

\bibitem[{Hashemi and Hwa(2016)}]{hashemi2016evaluation}
Homa~B Hashemi and Rebecca Hwa. 2016.
\newblock An evaluation of parser robustness for ungrammatical sentences.
\newblock In \emph{Proceedings of EMNLP}, pages 1765--1774.

\bibitem[{He et~al.(2016)He, Zhang, Ren, and Sun}]{he2016deep}
Kaiming He, Xiangyu Zhang, Shaoqing Ren, and Jian Sun. 2016.
\newblock Deep residual learning for image recognition.
\newblock In \emph{Proceedings of CVPR}, pages 770--778.

\bibitem[{Honnibal and Johnson(2014)}]{honnibal2014joint}
Matthew Honnibal and Mark Johnson. 2014.
\newblock Joint incremental disfluency detection and dependency parsing.
\newblock \emph{TACL}, 2:131--142.

\bibitem[{Junczys-Dowmunt et~al.(2018)Junczys-Dowmunt, Grundkiewicz, Guha, and
  Heafield}]{junczys2018approaching}
Marcin Junczys-Dowmunt, Roman Grundkiewicz, Shubha Guha, and Kenneth Heafield.
  2018.
\newblock Approaching neural grammatical error correction as a low-resource
  machine translation task.
\newblock In \emph{Proceedings of NAACL-HLT}, pages 595--606.

\bibitem[{Kaneko and Komachi(2019)}]{kaneko2019multi}
Masahiro Kaneko and Mamoru Komachi. 2019.
\newblock Multi-head multi-layer attention to deep language representations for
  grammatical error detection.
\newblock \emph{Computing Research Repository}, pages 883--891.

\bibitem[{Kaneko et~al.(2020)Kaneko, Mita, Kiyono, Suzuki, and
  Inui}]{kaneko2020encoder}
Masahiro Kaneko, Masato Mita, Shun Kiyono, Jun Suzuki, and Kentaro Inui. 2020.
\newblock Encoder-decoder models can benefit from pre-trained masked language
  models in grammatical error correction.
\newblock In \emph{Proceedings of ACL}, pages 4248--4254.

\bibitem[{Katsumata and Komachi(2020)}]{katsumata2020stronger}
Satoru Katsumata and Mamoru Komachi. 2020.
\newblock Stronger baselines for grammatical error correction using a
  pretrained encoder-decoder model.
\newblock In \emph{Proceedings of AACL}, pages 827--832.

\bibitem[{Kingma and Ba(2014)}]{kingma2014adam}
Diederik~P Kingma and Jimmy Ba. 2014.
\newblock Adam: A method for stochastic optimization.
\newblock \emph{arXiv preprint arXiv:1412.6980}.

\bibitem[{Kipf and Welling(2017)}]{DBLP:conf/iclr/KipfW17}
Thomas~N. Kipf and Max Welling. 2017.
\newblock Semi-supervised classification with graph convolutional networks.
\newblock In \emph{Proceedings of ICLR}.

\bibitem[{Kiyono et~al.(2019)Kiyono, Suzuki, Mita, Mizumoto, and
  Inui}]{kiyono2019empirical}
Shun Kiyono, Jun Suzuki, Masato Mita, Tomoya Mizumoto, and Kentaro Inui. 2019.
\newblock An empirical study of incorporating pseudo data into grammatical
  error correction.
\newblock In \emph{Proceedings of EMNLP-IJCNLP}, pages 1236--1242.

\bibitem[{Lewis et~al.(2020)Lewis, Liu, Goyal, Ghazvininejad, Mohamed, Levy,
  Stoyanov, and Zettlemoyer}]{lewis2020bart}
Mike Lewis, Yinhan Liu, Naman Goyal, Marjan Ghazvininejad, Abdelrahman Mohamed,
  Omer Levy, Veselin Stoyanov, and Luke Zettlemoyer. 2020.
\newblock Bart: Denoising sequence-to-sequence pre-training for natural
  language generation, translation, and comprehension.
\newblock In \emph{Proceedings of ACL}, pages 7871--7880.

\bibitem[{Li et~al.(2022)Li, Parnow, and Zhao}]{li2022syntax}
Zuchao Li, Kevin Parnow, and Hai Zhao. 2022.
\newblock Incorporating rich syntax information in grammatical error
  correction.
\newblock \emph{Information Processing \& Management}, 59(3):102891.

\bibitem[{Lichtarge et~al.(2020)Lichtarge, Alberti, and
  Kumar}]{lichtarge-etal-2020-data}
Jared Lichtarge, Chris Alberti, and Shankar Kumar. 2020.
\newblock Data weighted training strategies for grammatical error correction.
\newblock \emph{TACL}, pages 634--646.

\bibitem[{Lichtarge et~al.(2019)Lichtarge, Alberti, Kumar, Shazeer, Parmar, and
  Tong}]{lichtarge2019corpora}
Jared Lichtarge, Chris Alberti, Shankar Kumar, Noam Shazeer, Niki Parmar, and
  Simon Tong. 2019.
\newblock Corpora generation for grammatical error correction.
\newblock In \emph{Proceedings of NAACL-HLT}, pages 3291--3301.

\bibitem[{Marcinkiewicz(1994)}]{marcinkiewicz1994building}
Mary~Ann Marcinkiewicz. 1994.
\newblock Building a large annotated corpus of {English}: The penn treebank.
\newblock \emph{Using Large Corpora}, page 273.

\bibitem[{McClosky et~al.(2006)McClosky, Charniak, and
  Johnson}]{DBLP:conf/naacl/McCloskyCJ06}
David McClosky, Eugene Charniak, and Mark Johnson. 2006.
\newblock Effective self-training for parsing.
\newblock In \emph{Proceedings of NAACL-HLT}.

\bibitem[{Mita et~al.(2020)Mita, Kiyono, Kaneko, Suzuki, and
  Inui}]{mita2020self}
Masato Mita, Shun Kiyono, Masahiro Kaneko, Jun Suzuki, and Kentaro Inui. 2020.
\newblock A self-refinement strategy for noise reduction in grammatical error
  correction.
\newblock In \emph{Proceedings of EMNLP (Findings)}, pages 267--280.

\bibitem[{Nagata and Sakaguchi(2016)}]{nagata2016phrase}
Ryo Nagata and Keisuke Sakaguchi. 2016.
\newblock Phrase structure annotation and parsing for learner {English}.
\newblock In \emph{Proceedings of ACL}, pages 1837--1847.

\bibitem[{Nair and Hinton(2010)}]{nair2010rectified}
Vinod Nair and Geoffrey~E Hinton. 2010.
\newblock Rectified linear units improve restricted boltzmann machines.
\newblock In \emph{Proceedings of ICML}, pages 807--814.

\bibitem[{Napoles et~al.(2015)Napoles, Sakaguchi, Post, and
  Tetreault}]{napoles2015ground}
Courtney Napoles, Keisuke Sakaguchi, Matt Post, and Joel Tetreault. 2015.
\newblock Ground truth for grammatical error correction metrics.
\newblock In \emph{Proceedings of ACL (short)}, pages 588--593.

\bibitem[{Napoles et~al.(2017)Napoles, Sakaguchi, and
  Tetreault}]{napoles2017jfleg}
Courtney Napoles, Keisuke Sakaguchi, and Joel Tetreault. 2017.
\newblock {JFLEG}: A fluency corpus and benchmark for grammatical error
  correction.
\newblock In \emph{Proceedings of EACL}, pages 229--234.

\bibitem[{Ng et~al.(2014)Ng, Wu, Briscoe, Hadiwinoto, Susanto, and
  Bryant}]{ng2014conll}
Hwee~Tou Ng, Siew~Mei Wu, Ted Briscoe, Christian Hadiwinoto, Raymond~Hendy
  Susanto, and Christopher Bryant. 2014.
\newblock The {CoNLL}-2014 shared task on grammatical error correction.
\newblock In \emph{Proceedings of {CoNLL}: Shared Task}, pages 1--14.

\bibitem[{Omelianchuk et~al.(2020)Omelianchuk, Atrasevych, Chernodub, and
  Skurzhanskyi}]{omelianchuk2020gector}
Kostiantyn Omelianchuk, Vitaliy Atrasevych, Artem Chernodub, and Oleksandr
  Skurzhanskyi. 2020.
\newblock Gector--grammatical error correction: Tag, not rewrite.
\newblock In \emph{Proceedings of {BEA}@ACL}, pages 163--170.

\bibitem[{Ott et~al.(2019)Ott, Edunov, Baevski, Fan, Gross, Ng, Grangier, and
  Auli}]{ott2019fairseq}
Myle Ott, Sergey Edunov, Alexei Baevski, Angela Fan, Sam Gross, Nathan Ng,
  David Grangier, and Michael Auli. 2019.
\newblock fairseq: A fast, extensible toolkit for sequence modeling.
\newblock In \emph{Proceedings of NAACL-HLT(Demo)}, pages 48--53.

\bibitem[{Petrov and McDonald(2012)}]{petrov2012overview}
Slav Petrov and Ryan McDonald. 2012.
\newblock Overview of the 2012 shared task on parsing the web.
\newblock In \emph{First Workshop on Syntactic Analysis of Non-Canonical
  Language (SANCL)}.

\bibitem[{Pettersson et~al.(2012)Pettersson, Megyesi, and
  Nivre}]{pettersson2012parsing}
Eva Pettersson, Be{\'a}ta Megyesi, and Joakim Nivre. 2012.
\newblock Parsing the past-identification of verb constructions in historical
  text.
\newblock In \emph{Proceedings of LaTeCH@EACL}.

\bibitem[{Raffel et~al.(2020)Raffel, Shazeer, Roberts, Lee, Narang, Matena,
  Zhou, Li, and Liu}]{raffel2020exploring}
Colin Raffel, Noam Shazeer, Adam Roberts, Katherine Lee, Sharan Narang, Michael
  Matena, Yanqi Zhou, Wei Li, and Peter~J Liu. 2020.
\newblock Exploring the limits of transfer learning with a unified text-to-text
  transformer.
\newblock \emph{Journal of Machine Learning Research (JMLR)}, 21(140):1--67.

\bibitem[{Rei and Yannakoudakis(2016)}]{rei-yannakoudakis-2016-compositional}
Marek Rei and Helen Yannakoudakis. 2016.
\newblock Compositional sequence labeling models for error detection in learner
  writing.
\newblock In \emph{Proceedings of ACL}, pages 1181--1191.

\bibitem[{Rothe et~al.(2021)Rothe, Mallinson, Malmi, Krause, and
  Severyn}]{rothe2021recipe}
Sascha Rothe, Jonathan Mallinson, Eric Malmi, Sebastian Krause, and Aliaksei
  Severyn. 2021.
\newblock A simple recipe for multilingual grammatical error correction.
\newblock In \emph{Proceedings of ACL-IJCNLP}, pages 702--707.

\bibitem[{Sachan et~al.(2021)Sachan, Zhang, Qi, and
  Hamilton}]{sachan2021syntax}
Devendra Sachan, Yuhao Zhang, Peng Qi, and William~L Hamilton. 2021.
\newblock Do syntax trees help pre-trained transformers extract information?
\newblock In \emph{Proceedings of EACL}, pages 2647--2661.

\bibitem[{Sennrich et~al.(2016)Sennrich, Haddow, and
  Birch}]{sennrich-etal-2016-neural}
Rico Sennrich, Barry Haddow, and Alexandra Birch. 2016.
\newblock Neural machine translation of rare words with subword units.
\newblock In \emph{Proceedings of ACL}, pages 1715--1725.

\bibitem[{Shao et~al.(2021)Shao, Geng, Liu, Dai, Yang, Zhe, Bao, and
  Qiu}]{shao2021cpt}
Yunfan Shao, Zhichao Geng, Yitao Liu, Junqi Dai, Fei Yang, Li~Zhe, Hujun Bao,
  and Xipeng Qiu. 2021.
\newblock {CPT}: A pre-trained unbalanced transformer for both {Chinese}
  language understanding and generation.
\newblock \emph{arXiv preprint arXiv:2109.05729}.

\bibitem[{Stahlberg and Kumar(2021)}]{stahlberg2021synthetic}
Felix Stahlberg and Shankar Kumar. 2021.
\newblock Synthetic data generation for grammatical error correction with
  tagged corruption models.
\newblock In \emph{Proceedings of {BEA}@EACL}, pages 37--47.

\bibitem[{Sun et~al.(2022)Sun, Ge, Ma, Li, Wei, and Wang}]{sun2022unified}
Xin Sun, Tao Ge, Shuming Ma, Jingjing Li, Furu Wei, and Houfeng Wang. 2022.
\newblock A unified strategy for multilingual grammatical error correction with
  pre-trained cross-lingual language model.
\newblock \emph{arXiv preprint arXiv:2201.10707}.

\bibitem[{Sun et~al.(2021)Sun, Ge, Wei, and Wang}]{DBLP:conf/acl/SunGWW20}
Xin Sun, Tao Ge, Furu Wei, and Houfeng Wang. 2021.
\newblock Instantaneous grammatical error correction with shallow aggressive
  decoding.
\newblock In \emph{Proceedings of ACL-IJCNLP}, pages 5937--5947.

\bibitem[{Szegedy et~al.(2016)Szegedy, Vanhoucke, Ioffe, Shlens, and
  Wojna}]{szegedy2016rethinking}
Christian Szegedy, Vincent Vanhoucke, Sergey Ioffe, Jon Shlens, and Zbigniew
  Wojna. 2016.
\newblock Rethinking the inception architecture for computer vision.
\newblock In \emph{Proceedings of ICCV}, pages 2818--2826.

\bibitem[{Vaswani et~al.(2017)Vaswani, Shazeer, Parmar, Uszkoreit, Jones,
  Gomez, Kaiser, and Polosukhin}]{vaswani2017attention}
Ashish Vaswani, Noam Shazeer, Niki Parmar, Jakob Uszkoreit, Llion Jones,
  Aidan~N Gomez, {\L}ukasz Kaiser, and Illia Polosukhin. 2017.
\newblock Attention is all you need.
\newblock In \emph{Proceedings of NIPS}, pages 5998--6008.

\bibitem[{Velickovic et~al.(2018)Velickovic, Cucurull, Casanova, Romero,
  Li{\`{o}}, and Bengio}]{DBLP:conf/iclr/VelickovicCCRLB18}
Petar Velickovic, Guillem Cucurull, Arantxa Casanova, Adriana Romero, Pietro
  Li{\`{o}}, and Yoshua Bengio. 2018.
\newblock Graph attention networks.
\newblock In \emph{Proceedings of ICLR}.

\bibitem[{Wagner(2012)}]{wagner2012detecting}
Joachim Wagner. 2012.
\newblock \emph{Detecting grammatical errors with treebank-induced,
  probabilistic parsers}.
\newblock Ph.D. thesis, Dublin City University.

\bibitem[{Wan and Wan(2021)}]{wan2021syntax}
Zhaohong Wan and Xiaojun Wan. 2021.
\newblock A syntax-guided grammatical error correction model with dependency
  tree correction.
\newblock \emph{arXiv preprint arXiv:2111.03294}.

\bibitem[{Wang et~al.(2018)Wang, Pham, Yin, and Neubig}]{wang2018tree}
Xinyi Wang, Hieu Pham, Pengcheng Yin, and Graham Neubig. 2018.
\newblock A tree-based decoder for neural machine translation.
\newblock In \emph{Proceedings of EMNLP}, pages 4772--4777.

\bibitem[{Wang et~al.(2021)Wang, Wang, Dang, Liu, and
  Liu}]{wang2021comprehensive}
Yu~Wang, Yuelin Wang, Kai Dang, Jie Liu, and Zhuo Liu. 2021.
\newblock A comprehensive survey of grammatical error correction.
\newblock \emph{ACM Transactions on Intelligent Systems and Technology (TIST)},
  pages 1--51.

\bibitem[{Williams and Zipser(1989)}]{williams1989learning}
Ronald~J Williams and David Zipser. 1989.
\newblock A learning algorithm for continually running fully recurrent neural
  networks.
\newblock \emph{Neural computation}, 1(2):270--280.

\bibitem[{Wiseman and Rush(2016)}]{DBLP:conf/emnlp/WisemanR16}
Sam Wiseman and Alexander~M. Rush. 2016.
\newblock Sequence-to-sequence learning as beam-search optimization.
\newblock In \emph{Proceedings of EMNLP}, pages 1296--1306.

\bibitem[{Wolf et~al.(2020)Wolf, Debut, Sanh, Chaumond, Delangue, Moi, Cistac,
  Rault, Louf, Funtowicz, Davison, Shleifer, von Platen, Ma, Jernite, Plu, Xu,
  Scao, Gugger, Drame, Lhoest, and Rush}]{wolf-etal-2020-transformers}
Thomas Wolf, Lysandre Debut, Victor Sanh, Julien Chaumond, Clement Delangue,
  Anthony Moi, Pierric Cistac, Tim Rault, Rémi Louf, Morgan Funtowicz, Joe
  Davison, Sam Shleifer, Patrick von Platen, Clara Ma, Yacine Jernite, Julien
  Plu, Canwen Xu, Teven~Le Scao, Sylvain Gugger, Mariama Drame, Quentin Lhoest,
  and Alexander~M. Rush. 2020.
\newblock Transformers: State-of-the-art natural language processing.
\newblock In \emph{Proceedings of EMNLP (Demo)}, pages 38--45.

\bibitem[{Xia et~al.(2019)Xia, Li, Zhang, Zhang, Fu, Wang, and
  Si}]{xia2019syntax}
Qingrong Xia, Zhenghua Li, Min Zhang, Meishan Zhang, Guohong Fu, Rui Wang, and
  Luo Si. 2019.
\newblock Syntax-aware neural semantic role labeling.
\newblock In \emph{Proceedings of AAAI}, pages 7305--7313.

\bibitem[{Yan et~al.(2020)Yan, Qiu, and Huang}]{DBLP:journals/tacl/YanQH20}
Hang Yan, Xipeng Qiu, and Xuanjing Huang. 2020.
\newblock A graph-based model for joint {Chinese} word segmentation and
  dependency parsing.
\newblock \emph{TACL}, pages 78--92.

\bibitem[{Yannakoudakis et~al.(2011)Yannakoudakis, Briscoe, and
  Medlock}]{yannakoudakis2011new}
Helen Yannakoudakis, Ted Briscoe, and Ben Medlock. 2011.
\newblock A new dataset and method for automatically grading {ESOL} texts.
\newblock In \emph{Proceedings of ACL}, pages 180--189.

\bibitem[{Yasunaga et~al.(2021)Yasunaga, Leskovec, and
  Liang}]{DBLP:conf/emnlp/YasunagaLL21}
Michihiro Yasunaga, Jure Leskovec, and Percy Liang. 2021.
\newblock Lm-critic: Language models for unsupervised grammatical error
  correction.
\newblock In \emph{Proceedings of EMNLP}, pages 7752--7763.

\bibitem[{Yuan and Briscoe(2016)}]{yuan2016grammatical}
Zheng Yuan and Ted Briscoe. 2016.
\newblock Grammatical error correction using neural machine translation.
\newblock In \emph{Proceedings of NAACL-HLT}, pages 380--386.

\bibitem[{Yuan et~al.(2021)Yuan, Taslimipoor, Davis, and
  Bryant}]{DBLP:conf/emnlp/YuanTD021}
Zheng Yuan, Shiva Taslimipoor, Christopher Davis, and Christopher Bryant. 2021.
\newblock Multi-class grammatical error detection for correction: {A} tale of
  two systems.
\newblock In \emph{Proceedings of EMNLP}, pages 8722--8736.

\bibitem[{Zhang(2009)}]{zhang2009hsk}
Baolin Zhang. 2009.
\newblock Features and functions of the {HSK} dynamic composition corpus.
\newblock \emph{International {Chinese} Language Education}, 4:71--79.

\bibitem[{Zhang et~al.(2020{\natexlab{a}})Zhang, Zhang, Wang, Li, and
  Zhang}]{DBLP:conf/acl/ZhangZWLZ20}
Bo~Zhang, Yue Zhang, Rui Wang, Zhenghua Li, and Min Zhang. 2020{\natexlab{a}}.
\newblock Syntax-aware opinion role labeling with dependency graph
  convolutional networks.
\newblock In \emph{Proceedings of ACL}, pages 3249--3258.

\bibitem[{Zhang et~al.(2019)Zhang, Li, Fu, and
  Zhang}]{DBLP:conf/naacl/ZhangLFZ19}
Meishan Zhang, Zhenghua Li, Guohong Fu, and Min Zhang. 2019.
\newblock Syntax-enhanced neural machine translation with syntax-aware word
  representations.
\newblock In \emph{Proceedings of NAACL-HLT}, pages 1151--1161.

\bibitem[{Zhang et~al.(2020{\natexlab{b}})Zhang, Li, and
  Zhang}]{DBLP:conf/acl/ZhangLZ20}
Yu~Zhang, Zhenghua Li, and Min Zhang. 2020{\natexlab{b}}.
\newblock Efficient second-order treecrf for neural dependency parsing.
\newblock In \emph{Proceedings of ACL}, pages 3295--3305.

\bibitem[{Zhang et~al.(2022)Zhang, Li, Bao, Li, Zhang, Li, Huang, and
  Zhang}]{zhang2022mucgec}
Yue Zhang, Zhenghua Li, Zuyi Bao, Jiacheng Li, Bo~Zhang, Chen Li, Fei Huang,
  and Min Zhang. 2022.
\newblock {MuCGEC}: {a} multi-reference multi-source evaluation dataset for
  {Chinese} grammatical error correction.
\newblock In \emph{Proceedings of NAACL-HLT}, pages 3118--3130.

\bibitem[{Zhao et~al.(2018)Zhao, Jiang, Sun, and Wan}]{zhao2018overview}
Yuanyuan Zhao, Nan Jiang, Weiwei Sun, and Xiaojun Wan. 2018.
\newblock Overview of the {NLPCC} 2018 shared task: Grammatical error
  correction.
\newblock In \emph{CCF International Conference on Natural Language Processing
  and {Chinese} Computing (NLPCC)}, pages 439--445.

\bibitem[{Zhao and Wang(2020)}]{zhao2020maskgec}
Zewei Zhao and Houfeng Wang. 2020.
\newblock {MaskGEC}: Improving neural grammatical error correction via dynamic
  masking.
\newblock In \emph{Proceedings of AAAI}, pages 1226--1233.

\end{thebibliography}
\bibliographystyle{acl_natbib}

\appendix

\vspace{+1ex}
\begin{center}
\Large \textbf{Appendices} 
\end{center}

\vspace{+1ex}

\section{Hyper-parameters}
The main hyper-parameters adopted by SynGEC are presented in Table \ref{tab:hp}. When not using PLMs, the total training time is about 3 hours. When using PLMs, the training costs about 7 hours.
For fine-tuning BART on GEC data, we directly utilize the same hyper-parameters described in \citet{katsumata2020stronger}. When confronting sentences longer than the max input length, we keep them unchanged during predicting.
\label{sec:appendix:hp}
\begin{table}[h!]
\centering
\scalebox{0.74}{
\begin{tabular}{lc}
\hline
\textbf{Configuration}      & \textbf{Value}                        \\ \hline
\multicolumn{2}{c}{\textbf{Pre-training}}                  \\ \hline
Base architecture   & \begin{tabular}[c]{@{}l@{}}Transformer-base (w/o PLM)\\ Transformer-large (w/ PLM) \end{tabular}                           \\
Pretrained Language model   & BART-large \citep{lewis2020bart}                      \\
Number of epochs   & 60                           \\
Devices            & 8 Tesla V100 GPU (32GB)      \\
Batch size per GPU & 8096 tokens                        \\
Optimizer          & \begin{tabular}[c]{@{}c@{}}Adam \citep{kingma2014adam}\\ ($\beta_1=0.9,\beta_2=0.98,\epsilon=1 \times 10^{-8}$) \end{tabular}                        \\
Learning rate      &  $5 \times 10^{-4}$                          \\
Warmup updates             & 4000                         \\
Max source length  & 64 (English); 128 (Chinese)                           \\
Number of DepGCN layers  & 3                           \\
Dual context aggregation $\beta$ & 0.5 \\
Loss function      & \begin{tabular}[c]{@{}c@{}}Label smoothed cross entropy \\ (label-smoothing=0.1)\\\citep{szegedy2016rethinking} \end{tabular}  \\
Dropout            & 0.1 (w/o PLM); 0.3 (w/ PLM)                          \\ 
Dropout-src            & 0.2                          \\ 
\hline
\multicolumn{2}{c}{\textbf{Fine-tuning}}                   \\ \hline
Learning rate      & $5 \times 10^{-5}$                         \\
Warmup updates             & 1000                         \\
\hline
\multicolumn{2}{c}{\textbf{Generation}}                    \\ \hline
Beam size          & 12                           \\
Max input length  & 64 (English); 128 (Chinese)                           \\
\hline
\end{tabular}
}
\caption{Hyper-parameter values used in our experiments.}
\label{tab:hp}
\end{table}

\section{Experiments on JFLEG}
JFLEG \citep{napoles2017jfleg} is an English GEC evaluation dataset which focuses on fluency and uses the GLUE score \citep{napoles2015ground} as the evaluation metric. We evaluate the baseline and the SynGEC approach in Table \ref{tab:main:results} on JFLEG. Since JFLEG's scale is relatively small (only 747 sentences), we choose to present the results in the appendix. From Table \ref{tab:jfleg}, we can see that the syntactic knowledge still continuously improves the GEC performance over baselines with/without PLMs.
\label{sec:appendix:jfleg}

\begin{table}[h!]
\centering
\begin{tabular}{ccccc}
\hline
& \textbf{PLM} & \textbf{Syntax} & \textbf{GLUE}  & $\Delta$ \\ \hline
   \textbf{Baseline} &    \xmark         &   \xmark           & 58.15  & -        \\
      \textbf{SynGEC} &      \xmark        &      \cmark       & 60.14 & +1.99 \\ \hline
    \textbf{Baseline} &        \cmark       &     \xmark          & 61.53  & -        \\
      \textbf{SynGEC} &      \cmark       &       \cmark       & 62.15& +0.62 \\ \hline
\end{tabular}
\caption{The GLUE scores of different models on JFLEG benchmark.}
\label{tab:jfleg}
\end{table}

\section{The GED ability of GOPar}
\label{sec:appendix:ged}
We evaluate the binary Grammatical Error Detection (GED) performance of GOPar on two mainstream GED dataset, i.e., BEA-19-dev \citep{bryant2019bea} and FCE-test \citep{yannakoudakis2011new}. We follow \citet{rei-yannakoudakis-2016-compositional} and report token-level P/R/F values for detecting incorrect labels. 
Table \ref{tab:ged} shows the performance of GOPar and other leading GED models. 
When using the same training data, GOPar has a superior ability to detect grammatical errors. This phenomenon is very interesting and worthy of more in-depth study.
\begin{table}[h!]
\scalebox{0.67}{

\begin{tabular}{llcc}
\hline
\textbf{} & \textbf{}          & \textbf{BEA-19-\textit{dev}} & \textbf{FCE-\textit{test}} \\
\textbf{} & \textbf{Train-set} & \textbf{$\mbox{\textbf{F}}_{0.5}$}       & \textbf{$\mbox{\textbf{F}}_{0.5}$}     \\ \hline
\citet{bell-etal-2019-context}          & FCE-train                & 48.50               & 57.28             \\
\citet{kaneko2019multi}         & FCE-train                & --                  & 61.65             \\
\citet{DBLP:conf/emnlp/YuanTD021}         & FCE-train                & 65.54               & 72.93             \\
GOPar      & FCE-train                & \textbf{66.32}      & \textbf{74.10}    \\ \hline
GOPar      & CLang8             & \textbf{72.13}      & \textbf{71.53}    \\ \hline
\end{tabular}
}
\caption{Binary GED performance.}
\label{tab:ged}
\end{table}

\section{Download links of PLMs}
The download links of PLMs used in our experiments are listed below. We employ ELECTRA \citep{DBLP:conf/iclr/ClarkLLM20, cui2020revisiting} to build GOPar and BART \citep{lewis2020bart, shao2021cpt} to enhance our GEC model.
\begin{sloppypar}
 \begin{itemize}
     \item \href{https://huggingface.co/google/electra-large-discriminator}{ELECTRA-English-Large}.
     \item \href{https://huggingface.co/hfl/chinese-electra-180g-large-discriminator}{ELECTRA-Chinese-Large}.
     \item \href{https://huggingface.co/facebook/bart-large}{BART-English-Large}.
     \item \href{https://huggingface.co/fnlp/bart-large-chinese}{BART-Chinese-Large}.
 \end{itemize}
\end{sloppypar}

\label{sec:appendix:download}

\end{document}